\documentclass[sigconf, nonacm]{acmart}

\usepackage{algorithm}
\usepackage{algpseudocode}
\usepackage{booktabs}
\usepackage{multirow} 
\usepackage{graphicx}
\usepackage{subcaption} 
\usepackage{enumitem}

\newcommand\vldbdoi{XX.XX/XXX.XX}

\newcommand\vldbavailabilityurl{URL_TO_YOUR_ARTIFACTS}
\newcommand\vldbpagestyle{empty} 

\begin{document}
\title{HeteroHub: An Applicable Data Management Framework for Heterogeneous Multi-Embodied Agent System}


\author{Xujia Li}
\email{leexujia@ust.hk}
\affiliation{%
  \institution{HKUST}
  \authornote{HKUST is the abbreviation for The Hong Kong University of Science and Technology.}
  \city{Hong Kong SAR}
  \country{China}
}

\author{Xin Li}
\email{xli494@connect.hkust-gz.edu.cn}
\affiliation{%
  \institution{HKUST (GZ)}
  \city{GuangZhou}
  \country{China}
}

\author{Junquan Huang}
\email{junquan@m.scnu.edu.cn}
\affiliation{%
  \institution{HKUST (GZ)}
  \city{GuangZhou}
  \country{China}
}

\author{Beirong Cui}
\email{bcui472@connect.hkust-gz.edu.cn}
\affiliation{%
  \institution{HKUST (GZ)}
  \city{GuangZhou}
  \country{China}
}

\author{Zibin Wu}
\email{zwu945@connect.hkust-gz.edu.cn}
\affiliation{%
  \institution{HKUST (GZ)}
  \city{GuangZhou}
  \country{China}
}

\author{Lei Chen}
\email{leichen@cse.ust.hk}
\affiliation{%
  \institution{HKUST (GZ) \& HKUST}
  \city{GuangZhou}
  \country{China}
}

\begin{abstract}
Heterogeneous Multi-Embodied Agent Systems involve coordinating multiple embodied agents with diverse capabilities to accomplish tasks in dynamic environments. This process requires the collection, generation, and consumption of massive, heterogeneous data, which primarily falls into three categories: static knowledge regarding the agents, tasks, and environments; multimodal training datasets tailored for various AI models; and high-frequency sensor streams. However, existing frameworks lack a unified data management infrastructure to support the real-world deployment of such systems. To address this gap, we present \textbf{HeteroHub}, a data-centric framework that integrates static metadata, task-aligned training corpora, and real-time data streams. The framework supports task-aware model training, context-sensitive execution, and closed-loop control driven by real-world feedback. In our demonstration, HeteroHub successfully coordinates multiple embodied AI agents to execute complex tasks, illustrating how a robust data management framework can enable scalable, maintainable, and evolvable embodied AI systems. 
\end{abstract}

\maketitle

\pagestyle{\vldbpagestyle}




\section{Introduction}
Heterogeneous Multi-Embodied Agent Systems are poised to become integral to daily life such as buying coffee, collecting deliveries, cleaning homes, and conducting security patrols \cite{emos,COHERENT}. However, enabling seamless collaboration among diverse hardware devices while coordinating various task-specific AI models creates a critical demand for a robust, applicable data management framework. Specifically, managing the vast amounts of heterogeneous data raises significant challenges \cite{magnus,eckmann2026vision}. The related data spans three primary categories: static knowledge regarding agent profiles and task descriptions, dynamic sensor streams, and adequate training datasets for AI models. Beyond handling diverse data types, this effective data management framework also supports the entire operational lifecycle: maintaining static metadata, facilitating AI model training, and processing real-time data during task execution \cite{wang2024optimizing}.

In this demo, we present HeteroHub, a comprehensive data management solution designed to orchestrate the lifecycle of data in embodied intelligent systems. HeteroHub is composed of three interconnected layers: (1) \textbf{Static Knowledge Management}, which includes detailed profiles of agents, a task graph representing executable workflows, a model library housing AI algorithms, and environment information; (2) \textbf{Training Data Fabric}, which focuses on collecting and organizing multimodal datasets for training perception, reasoning, and interaction models; and (3) \textbf{Real-Time Sensor Stream Management}, which handles the ingestion, processing, and routing of sensor data during task execution. At the core of HeteroHub is a task-aligned data structure that ensures all data elements are semantically linked to specific tasks. This alignment enables precise querying and retrieval of relevant data, facilitating efficient training and deployment of AI models. Another key innovation is its adaptability to new scenarios. When a new device, capability, or model is introduced, the system can generate synthetic workflows, augment the training set, and perform lightweight fine-tuning. This ensures that the system remains aligned with evolving operational requirements and learns from every interaction.

To demonstrate the effectiveness of HeteroHub, we deploy it in a smart \textbf{campus logistics scenario}. Specifically, we orchestrate heterogeneous embodied agents to collaborate on complex tasks, including cross-floor and indoor-outdoor navigation, as well as autonomous pickup and delivery. Our demonstration highlights how HeteroHub manages the entire data lifecycle from initial data collection and model training to real-time task planning and execution. By presenting HeteroHub, we aim to illustrate \textbf{the critical role of robust data management in enabling adaptable and scalable embodied AI systems}. We believe this work will inspire further research into data-centric approaches for embodied AI, fostering innovations that bridge the gap between theoretical advancements and practical applications. The \textbf{demo video} is available on \url{https://youtu.be/rXEKhaa7Wy0}.

\section{Framework Overview}

\subsection{Embodied Agent Static Information Hub}

In Fig.~\ref{fig1}, we present the Static Information Hub (SI-Hub), which is a structured data management framework designed to organize and serve the static knowledge required by heterogeneous embodied agent systems. SI-Hub consists of four interlinked modules: 

\begin{figure}[tbp]
    \centering
    \includegraphics[width=1\linewidth]{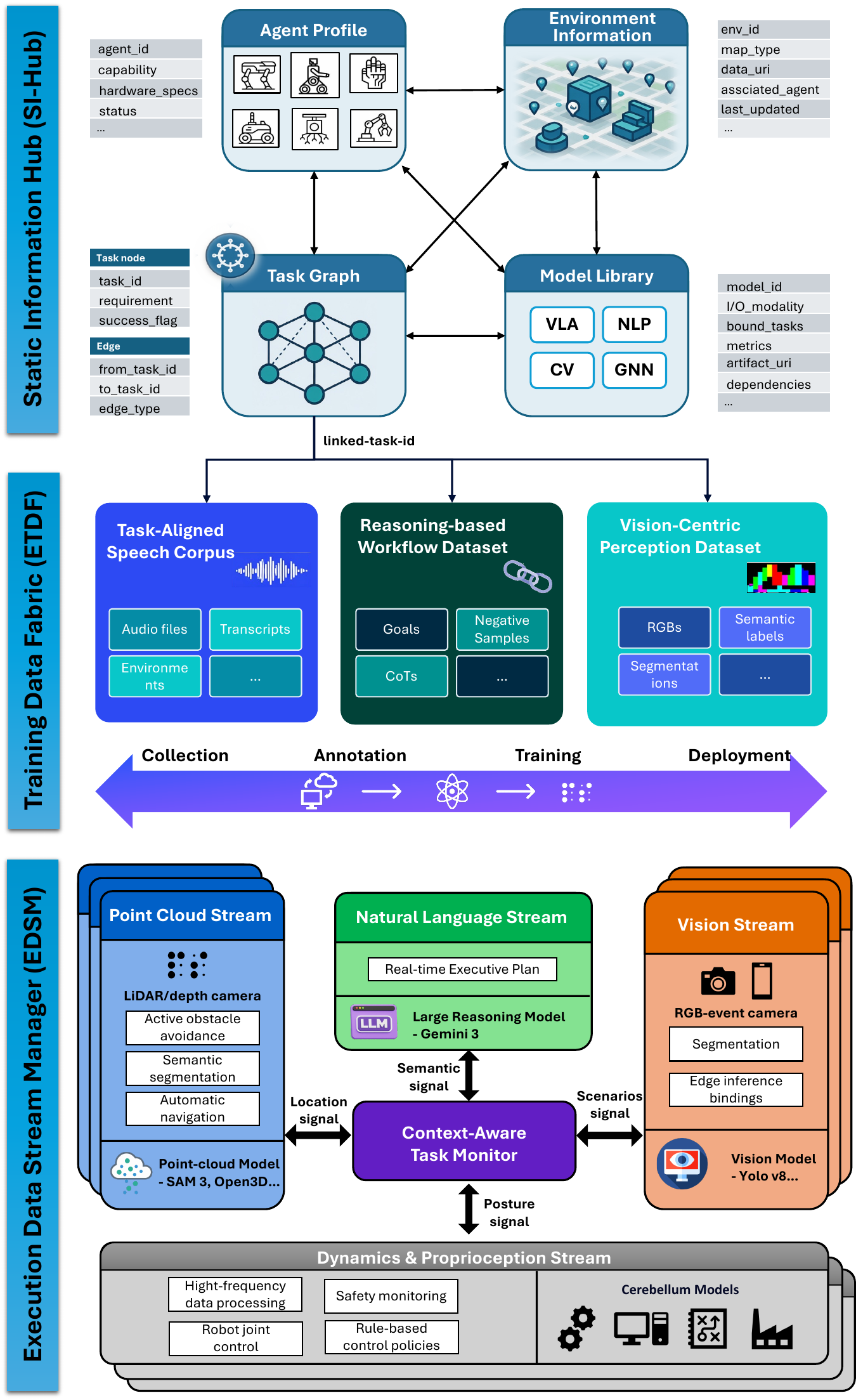}
    \caption{HeterHub for Multi-agent Embodied AI Systems.}
    \label{fig1}
\end{figure}

The \textbf{Agent Profiles} module maintains a registry of all embodied agents, capturing their physical specifications, such as degrees of freedom, sensor suites, functional capabilities, and supported tasks. Each agent is assigned a globally unique URI, allowing unambiguous reference across the system.

The \textbf{Task Graph} encodes the space of executable tasks as a directed property graph. Nodes represent atomic or composite tasks annotated with input and output schemas, required capabilities, and success criteria. Edges model sequential, parallel, or conditional dependencies between tasks. This graph enables high-level task planning and compatibility checks against agent capabilities.

The \textbf{Model Library} catalogs all AI models used for perception, planning, control, or language grounding. For each model, we store metadata including modality of input abnd output data, performance metrics, version, and the relations to one or more tasks in the Task Graph. Model artifacts themselves are stored externally with only URIs retained in the registry to ensure scalability.

Finally, the \textbf{Environment Information} module stores digital representations of operational environments, such as point clouds, occupancy grids, or semantic maps. Each environment entry includes metadata about static objects, dynamic zones, and associated agents, along with a reference to the actual map file. Environments are also linked to tasks commonly executed within them, facilitating context-aware deployment.

All four modules are interconnected through standardized URI-based references. For example, an agent profile references task IDs and a task node references model IDs. This design enables rich cross-module queries. For instance, "Which agents in environment $E_1$ can execute task $T_1$, and which models should be loaded?", while maintaining loose coupling for independent updates. SI-Hub thus provides a scalable, queryable, and extensible foundation for managing the static knowledge backbone of multi-agent embodied intelligence systems.

\subsection{Embodied AI Training Data Fabric (ETDF)}

ETDF is a multimodal data storage for the entire model training lifecycle, spanning data collection, annotation, training, and deployment. ETDF is grounded in the principle of \textbf{task-centric data curation}, which means every training sample is explicitly linked to a node in the Task Graph.

ETDF currently comprises three core components. The first module, the \textbf{Task-Aligned Speech Corpus}, manages voice interaction data collected in context-specific scenarios. Each entry consists of a recorded audio utterance, a corresponding human-verified transcript, and an intent representation (e.g., grasp(object="mug")). This corpus directly supports fine-tuning automatic speech recognition and natural language understanding models that map user commands to actionable system intents.

The second module, \textbf{Reasoning-Based Workflow Dataset}, captures high-level planning knowledge by pairing LLM-generated Chains of Thought (CoTs) with both flawless multi-agent trajectories and deliberately flawed plans \cite{pan2025database}. Each entry includes a natural-language goal, e.g., "Transport equipment to E4", a step-by-step reasoning trace, and machine-readable action sequences. To ensure strict adherence to physical and morphological constraints, its training leverages the Reasoning-Based Workflow Dataset rather than relying solely on successful demonstrations. Crucially, the rejected plans are injected with compound errors, such as a drone navigating indoors or a legged robot attempting to open a door. The training data consists of reasoning-augmented preference tuples: (task context, chosen flawless plan, rejected flawed plan, penalty score). These preference pairs are generated through our automated three-phase pipeline: (1) extracting localized subgraphs to provide high-density physical contexts without attention dilution; (2) probabilistic injection of compound errors, including capability mismatches and coordination failures, to construct highly deceptive negative samples; and (3) rigorous evaluation via a hybrid symbolic-semantic validator to compute a quantitative penalty score based on error severity, instead of simple binary success or failure labels.

The third module, the \textbf{Vision-Centric Perception Dataset}, stores annotated visual data, including RGB, semantic labels for training specific perception models. Annotations are all aligned with the object vocabulary defined in the corresponding task scope. Images and annotations are organized hierarchically by task and object class, and exported in standard formats compatible with mainstream computer vision frameworks \cite{Garcia2024TowardsGV}. Critically, camera intrinsics and scene context are preserved to support geometry-aware training and domain adaptation.




\subsection{Execution Data Stream Manager}

To close the perception–decision–action loop during task execution, we introduce the Execution Data Stream Manager (EDSM). Unlike traditional data logging systems that treat sensors as passive sources \cite{Robroek}, EDSM treats sensor data as task-driven semantic signals, dynamically activating processing pipelines based on the current executable plan derived from the high-level reasoning module.

EDSM manages three primary modalities. The \textbf{point cloud stream}, derived from 3D data collected by LiDARs or depth cameras, undergoes real-time preprocessing on the robot edge using the known environment map from the Static Knowledge Hub. Following the processing on the edge, the real-time status of agent, specifically location signals in this demo, is fed back to the central task monitor to verify the completion status of the current sub-task. Simultaneously, the point cloud data is processed on-device by SLAM algorithms to enable autonomous navigation and active obstacle avoidance for the embodied agent. Secondly, the \textbf{visual data stream} is transmitted to lightweight, task-specific trained vision models on the edge, e.g., YOLO for detection or DINOv2 for feature extraction. To optimize bandwidth, only frames containing scenario-relevant content trigger a upload to the central task monitor. Visual detections are fused with point cloud data to refine spatial understanding for more effective sub-task execution. For instance, by projecting 2D bounding boxes into 3D space to guide grasp planning—thereby enabling more effective sub-task execution. The \textbf{dynamics and proprioception data stream} captures high-frequency signals from joint encoders, IMUs, and force or torque sensors. This data is primarily processed in real-time on the edge by diverse cerebellar control models to perform state estimation and continuously monitor safety constraints, such as joint torque limits and balance stability. When an executable plan specifies a control policy, the EDSM dynamically invokes the corresponding cerebellum, feeding it real-time proprioceptive data with the task goal. The policy then generates low-level commands.

The central \textbf{Context-Aware Task Monitor} primarily aggregates and analyzes processed signals from various local data streams to assess the completion status of each sub-task. It aligns this progress with the real-time executive plan generated by the large reasoning model. If deviations or errors are detected, the monitor triggers the reasoning large model to regenerate the work execution plan. Upon receiving an updated executable plan, the Monitor redistributes sub-tasks to the local embodied agents.

\section{Demonstration Proposal}
This demo focuses on logistics transportation tasks within a campus and consists of two main parts. The first part involves using ETDF in HeteroHub to \textbf{train deep learning models} tailored to specific logistics tasks. The second part leverages the trained brain and cerebellum models to interact in real time with static information from SI-Hub and dynamic signals from EDSM, enabling embodied agents to \textbf{execute tasks} effectively. 


\begin{figure*}[t]
    \centering
    \includegraphics[width=1\linewidth]{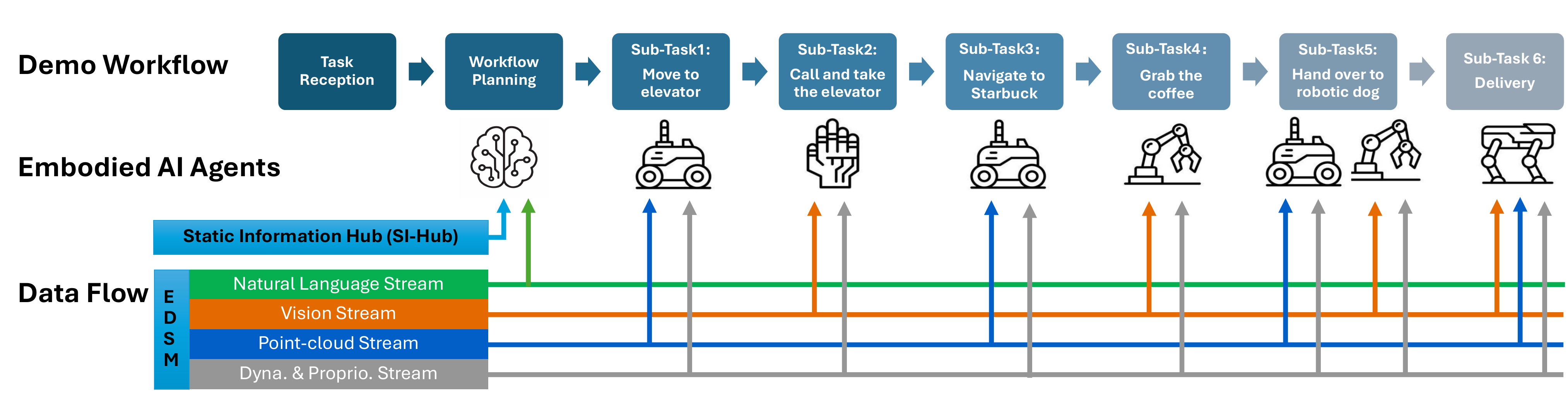}
    \caption{Orchestrating Multi-Agent Embodied AI: A Data-Driven Workflow for "Bring Me a Coffee from Starbucks"}
    \label{fig2}
\end{figure*}

\subsection{Model Training Demonstration}

We will demonstrate the data sample, training implementations, and training result of following models using HeteroHub.

\subsubsection{\textbf{Scenario 1: Training the Brain for Campus Logistics}}
The "brain", which is a fine-tuned large reasoning model, is responsible for translating abstract user goals, such as "Buy me a cup of coffee from Starbucks", into concrete execution plans. Training proceeds in two stages. First, supervised fine-tuning teaches the model to map spatial contexts and goals to executable plans using chain-of-thought prompting. This phase establishes the foundational reasoning steps required to generate valid structured outputs. Second, we apply Direct Preference Optimization (DPO) to refine the policy. This fine-grained penalty feedback directly enables DPO, where the alignment margin is scaled proportionally to the execution violation \cite{acm}. DPO utilizes the penalty scores computed by the hybrid validator to dynamically scale the alignment margin. This forces the optimization landscape to aggressively penalize severe physical violations, e.g., a drone navigating indoors, while applying softer corrections to minor coordination inefficiencies. By grounding every preference pair in explicit physical constraints, this module ensures the planner generates only feasible, execution-consistent strategies.


\subsubsection{\textbf{Scenario 2: Training the Cerebellums for Task-Specific Perception}}
In this demo, two cerebellum models are trained using ETDF, responsible for speech understanding and visual perception. 

\textbf{For speech}, the Task-Aligned Speech Corpus provides audio-transcript-intent triples annotated with task context. We train a compact Whisper variant for ASR and a BERT-based intent parser whose output space is dynamically constrained by the active task’s object vocabulary. This ensures the NLU module ignores out-of-scope commands, thereby improving robustness. 

\textbf{For vision}, the Vision-Centric Perception Dataset supplies RGB-Depth images and annotations organized by task and object class. Models like YOLOv8 or Segment Anything are fine-tuned exclusively on data relevant to active tasks. Multi-task learning is employed when object categories overlap across tasks. For example, we jointly train the policies for "pressing the elevator button" and "grasping a coffee cup" on the same robotic arm and camera suite, sharing a common backbone to improve sample efficiency. 


\subsection{Task Execution Demonstration}

To demonstrate the integration of HeteroHub with heterogeneous embodied agents, we show the full execution process for a representative task: "\textbf{Grab a coffee from the Starbucks}." This demo seamlessly orchestrates static knowledge, trained models, real-time sensing, and physical actuation through \textbf{five successive scenarios}.

\subsubsection{\textbf{Scenario 1: Task Reception and Workflow Planning}}
The interaction begins when a human issues a spoken command. The audio stream is captured by the microphone array and routed to the Task-Aligned Speech Corpus. The speech model transcribes the utterance and parses it into a structured intent. Because this is a natural-language goal rather than a low-level instruction, it is directly forwarded to the brain. Upon receiving the intent, the brain consults the Task Graph to decompose the goal into executable subtasks. Leveraging its training on reasoning-augmented workflows, it generates a context-aware plan grounded in the system’s capabilities. To ensure feasibility, each subtask is bonded to a specific agent and a model registered in the Model Library.

  
\subsubsection{\textbf{Scenario 2: Autonomous Navigation}} The Task Monitor dispatches the plan generated in Scenario 1 step-by-step. First, it activates the Chassis Agent $(agent://chassis/01)$ and instructs it to navigate to the elevator. During navigation, the chassis agent retrieves the static map $(env://5th\_floor)$ from SI-Hub and simultaneously performs real-time environmental perception using LiDAR and proprioceptive sensors managed by the EDSM module. It then uses a SLAM algorithm to localize and navigate to the destination of the first sub-task \cite{robotics11010024}.

\subsubsection{\textbf{Scenario 3: Vision-based Control}} 

Upon reaching the elevator, the Task Monitor invokes the visual perception model $(model://yolo\_elevator/01)$ deployed on the edge compute unit to perform semantic segmentation of the visual input. The Arm Agent then performs real-time detection of the elevator buttons on the wall using the live Vision Stream. Upon successful detection, the system triggers the next action: the mechanical arm $(agent://arm/02)$ is activated to press the designated elevator button. Similar vision-based control mechanisms also appear in other sub-tasks within this demo, including recognizing visual cues outside the elevator to help the chassis identify the current floor, and performing semantic segmentation on images of coffee bags to determine precise grasping points for the robotic arm. These sub-tasks require fine-grained control, which is achieved through trained cerebellum models in conjunction with the vision streams managed by EDSM.


\subsubsection{\textbf{Scenario 4: Collaboration among Multiple Agents}}
To demonstrate Hetero-Hub’s capability in supporting collaborative tasks among a group of embodied agents, our demo features a cooperation scenario involving the chassis agent, the arm agent, and a robot dog. They jointly transport a coffee bag through handover and ultimately deliver it to a confined office area. Leveraging the Task Monitor’s scheduling capability for both tasks and data streams, the three agents concurrently access different modalities of data from Hetero-Hub: point-cloud streams for navigation, and vision signals for robotic arm control and vision-based navigation in narrow spaces.



\subsubsection{\textbf{Scenario 5: Real-Time Feedback and Plan Refinement}}
After each subtask completes or fails, the agent sends a structured status report back to the brain. If a step fails, e.g., grasp unsuccessful due to slippage, the failure signal includes diagnostic metadata $(event://low\_force\_reading)$. The brain can then trigger a local retry, like reposition and re-grasp, or trigger a fall back to an alternative workflow branch defined in the Task Graph, e.g., "ask user for help". This closed-loop feedback ensures robustness while simultaneously enriching the Reasoning-Augmented Workflow Dataset with real-world experience, thereby closing the cycle between task execution and system continual learning.

\section{Conclusion}
HeteroHub demonstrates that robust embodied intelligence fundamentally relies on principled data management. By unifying static knowledge, training data, and real-time sensor streams within a task-centric framework, HeteroHub enables traceable, scalable, and adaptive robot systems. Our demo illustrates how this framework are not just supportive but essential to coordinating heterogeneous agents in dynamic environments. As embodied AI scales to real-world deployments, we hope this demo inspires deeper collaboration between the database and robotics communities.


\bibliographystyle{ACM-Reference-Format}
\bibliography{reference}

\end{document}